\documentclass{article}

\usepackage{microtype}
\usepackage{graphicx}
\usepackage{subcaption}
\usepackage{booktabs} 

\usepackage{hyperref}

\usepackage[preprint]{icml2026}

\usepackage{amsmath}
\usepackage{amssymb}
\usepackage{mathtools}
\usepackage{amsthm}
\usepackage{multirow}

\usepackage[capitalize,noabbrev]{cleveref}
\usepackage{listings}
\theoremstyle{plain}

\theoremstyle{definition}

\theoremstyle{remark}

\usepackage[textsize=tiny]{todonotes}
\usepackage{placeins}
\usepackage[table]{xcolor}

\definecolor{mygreen}{RGB}{197,224,180}
\definecolor{myyellow}{RGB}{255,230,153}
\definecolor{myred}{RGB}{255,171,171}
\definecolor{myorange}{RGB}{250,209,192}
\definecolor{mypurple}{RGB}{225,213,231}

\icmltitlerunning{Revisiting the Travel Planning Capabilities of Large Language Models}

\begin{document}

\twocolumn[
  \icmltitle{Revisiting the Travel Planning Capabilities of Large Language Models}



  \icmlsetsymbol{equal}{*}

  \begin{icmlauthorlist}
    \icmlauthor{Bo-Wen Zhang}{equal,lab,issch}
    \icmlauthor{Jin Ye}{equal,lab,issch}
    \icmlauthor{Peng-Yu Hua}{equal,lab,issch}
    \icmlauthor{Jia-Wei Cao}{equal,lab,issch}
    \icmlauthor{Jie-Jing Shao}{lab}
    \icmlauthor{Yu-Feng Li}{lab,aisch}
    \icmlauthor{Lan-Zhe Guo}{lab,issch}
  \end{icmlauthorlist}

  \icmlaffiliation{lab}{State Key Laboratory of Novel Software Technology, Nanjing University, China}
  \icmlaffiliation{issch}{School of Intelligence Science and Technology, Nanjing University, China}
  \icmlaffiliation{aisch}{School of Artificial Intelligence, Nanjing University, China}

  \icmlcorrespondingauthor{Lan-Zhe Guo}{guolz@nju.edu.cn}

  \icmlkeywords{}

  \vskip 0.3in
]



\printAffiliationsAndNotice{}  

\begin{abstract}
Travel planning serves as a critical task for long-horizon reasoning, exposing significant deficits in LLMs. However, existing benchmarks and evaluations primarily assess final plans in an end-to-end manner, which lacks interpretability and makes it difficult to analyze the root causes of failures. To bridge this gap, we decompose travel planning into five constituent atomic sub-capabilities, including \emph{Constraint Extraction}, \emph{Tool Use}, \emph{Plan Generation}, \emph{Error Identification}, and \emph{Error Correction}. We implement a decoupled evaluation protocol leveraging oracle intermediate contexts to rigorously isolate these components, thereby measuring the atomic performance boundary without the noise of cascading errors. Our results highlight a clear contrast in performance: while LLMs are proficient in extracting explicit constraints, they struggle to infer implicit, open-world requirements. Furthermore, they exhibit structural biases in plan generation and suffer from ineffective self-correction, characterized by excessive sensitivity and erroneous persistence. These findings offer precise directions for improving LLM reasoning and planning abilities.
\end{abstract}

\section{Introduction}
\begin{figure*}
    \centering
    \includegraphics[width=0.95\linewidth]{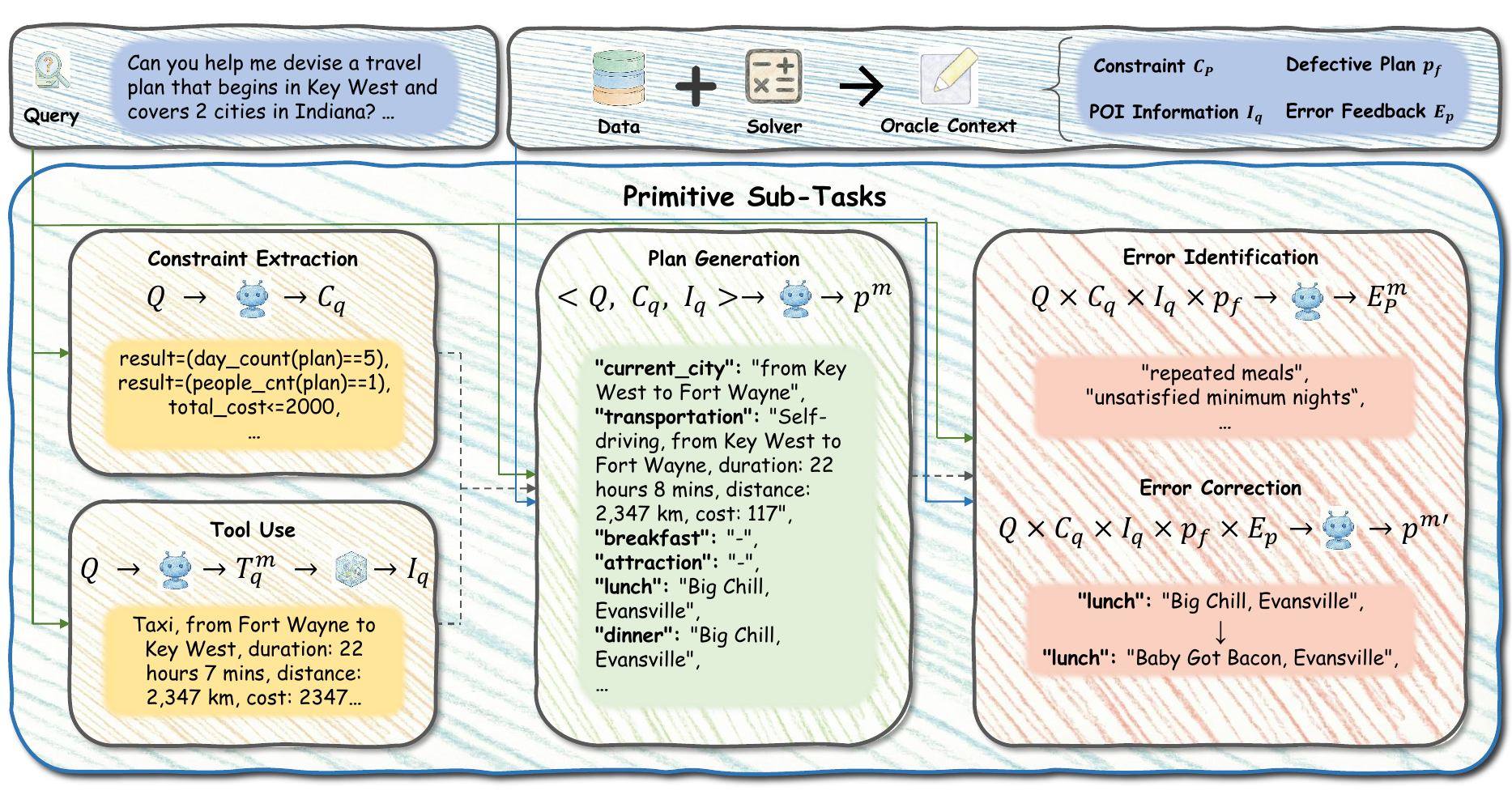}
    \caption{An overview of our decoupled evaluation protocol. We assess each atomic sub-capability independently using oracle intermediate contexts to prevent error propagation.}
    \label{fig:framework}
\end{figure*}

Large Language Models (LLMs) are increasingly recognized as a viable pathway toward Artificial General Intelligence (AGI)~\cite{zhao2023survey}. Beyond standard capabilities, recent advancements have demonstrated that LLMs possess preliminary reasoning abilities, achieving impressive results in logic-intensive domains such as mathematics~\cite{openai2024o1,guo2025deepseek}. By evolving into autonomous agents, they have further exhibited remarkable versatility across diverse tasks, ranging from web navigation to scientific simulation~\cite{scienceworld2022,jansen2024discoveryworld,li2025labutopia}. However, despite these promising results, LLMs still face significant challenges in complex reasoning and planning tasks that demand strict constraint satisfaction. Travel planning serves as a particularly good example of this limitation. Specifically, this task requires agents to orchestrate heterogeneous tools to synthesize itineraries that strictly adhere to complex user preferences, rigid hard constraints, and intricate spatiotemporal dependencies. Consequently, even advanced models struggle to reconcile these multi-dimensional requirements, with success rates often remaining low in realistic settings~\cite{xie2024travelplanner,shao2026chinatravel,chaudhuri2025tripcraft}.


Given the substantial research significance and practical applicability of the travel planning task, diverse evaluation benchmarks have been introduced, such as TravelPlanner~\cite{xie2024travelplanner}, ChinaTravel~\cite{shao2026chinatravel}, TripCraft~\cite{chaudhuri2025tripcraft}, etc. However, existing benchmarks predominantly rely on a coupled, end-to-end execution paradigm. The fundamental limitation of this approach is error propagation: a failure in an early stage (e.g., constraint extraction) inevitably corrupts the inputs for downstream stages, rendering the evaluation of subsequent capabilities (e.g., core planning) effectively meaningless. Consequently, the root causes of failure remain entangled. It becomes impossible to distinguish whether a failed itinerary stems from an actual deficit in combinatorial reasoning or simply from receiving incorrect constraints to begin with. This opacity hinders researchers from identifying the true bottlenecks, significantly preventing the targeted improvement of LLM and agent reasoning abilities.


To enable a fine-grained analysis of LLM capabilities and find clear pathways for targeted performance enhancement, we design a comprehensive diagnostic study to isolate reasoning bottlenecks. Specifically, we deconstruct the complex travel planning process into five atomic sub-tasks: Constraint Extraction, Tool Use, Plan Generation, Error Identification, and Plan Correction. Crucially, to circumvent error propagation, we implement a decoupled evaluation protocol utilizing oracle intermediate contexts. This approach allows us to assess the intrinsic capability of each module in isolation, thereby precisely pinpointing deficits and guiding future optimizations. Utilizing this diagnostic framework, our experiments reveal three critical observations regarding the component-wise capabilities of LLMs:

\begin{itemize}
    \item \textbf{Deficits in Open-World Constraint Inference:} While models demonstrate proficiency in extracting explicit, closed-world constraints and handling tool requirements, they falter significantly in open-world settings. Our analysis reveals that agents struggle to infer the implicit, common-sense constraints that are indispensable for realistic and executable planning.

    \item \textbf{Distribution-Induced Biases in Plan Generation:} In the core planning phase, agents exhibit structural generation biases in both the selection and temporal arrangement of Points of Interest (POIs). Rather than strictly optimizing for logical feasibility, generated itineraries often reflect distributional priors that diverge from the valid solutions identified by symbolic solvers.
    
    \item \textbf{Breakdown of Self-Reflection Mechanisms:} 
    We provide a detailed characterization of the reflection loop, identifying two critical failure modes: (i) \emph{over-sensitivity}, where valid steps are frequently misclassified as erroneous (false positives); and (ii) 
    \emph{erroneous persistence}, where agents fail to rectify identified errors even after multiple revision attempts, revealing a limited correction capability.
    
\end{itemize}

\section{Related Works}

\subsection{LLM-based Autonomous Agents}

LLM-based agents have emerged as powerful paradigms for tackling complex tasks by decomposing high-level instructions into executable actions. An LLM agent architecture typically comprises three core components: planning, memory, and tool use~\cite{weng2023agent}. Representative frameworks such as ReAct~\cite{yao2022react} and Reflexion~\cite{shinn2023reflexion} enhance the agent's reasoning by interleaving thought generation with action execution or incorporating verbal reinforcement. Meanwhile, systems like AutoGPT~\cite{yang2023auto} demonstrate the potential of agents in autonomous loop execution and tool orchestration. However, existing frameworks often struggle in complex tasks involving multiple, interdependent constraints, where satisfying one requirement may inadvertently violate another, leading to suboptimal or invalid plans.

\subsection{Travel Planning Benchmarks}
A surge of benchmarks has recently emerged in the domain of travel planning. They generally require LLMs to gather relevant information and generate itineraries that pass final validity checks. The underlying constraints typically fall into two categories: \textit{common-sense constraints}, which enforce universal consistency across all plans, and \textit{logical constraints}, which vary dynamically based on specific user requirements~\cite{xie2024travelplanner}. 
Regarding data sources, methodologies have transitioned from synthetic constraint sampling to LLM-based persona simulation~\cite{singh2024personal,chaudhuri2025tripcraft} and authentic real-world queries~\cite{shao2026chinatravel,qu2025tripscore}. 
In terms of plan granularity, benchmarks like ChinaTravel and TripCraft have further incorporated inter-POI transportation details to enable more precise scheduling. 
For handling soft preferences beyond hard constraints, evaluation approaches vary: some formulate preference adherence as an optimization problem for ranking~\cite{shao2026chinatravel}, others calculate semantic similarity scores~\cite{chaudhuri2025tripcraft}, while the majority employ LLMs or Reward Models as judges~\cite{singh2024personal,wang2025triptailor,shao2025personal}. 
Furthermore, features such as simulated user interactions~\cite{deng2025retail} and unexpected interruptions~\cite{karmakar2025triptide} have been introduced to enhance realism. 
However, despite these diversities, most existing works predominantly rely on holistic, outcome-based metrics (e.g., success rate). This coupled evaluation paradigm tends to entangle various capabilities, obscuring the specific reasoning bottlenecks that lead to failure.

\subsection{LLM for Travel Planning}

Current research to solve the travel planning tasks generally falls into four categories: (1) Neuro-Symbolic Integration: To ensure constraint satisfaction, several studies integrate LLMs with external symbolic solvers or verifiers~\cite{kambhampati2024position,hao2025large,ju-etal-2024-globe}. By translating natural language requirements into formal constraints, these neuro-symbolic methods achieve high performance in closed-world settings. (2) Task Decomposition \& Hierarchical Planning: To mimic human-like reasoning without external solvers, some efforts have focused on designing sophisticated planning pipelines. For example, HLRF~\cite{xie2024human} decomposes tasks into outline generation, information collection, and planning phases. Similarly, ISP~\cite{huainteractive} introduces approximation agents for hierarchical planning, and TGMA~\cite{deng2025retail} adopts a similar iterative refinement strategy to handle complex dependencies. (3) Multi-Agent Collaboration: Beyond single-agent pipelines, role specialization has emerged as a promising direction. Frameworks like PMC~\cite{zhang2025planning}, DPPM ~\cite{lu2025decompose}, and Atlas~\cite{choi2025atlas} distribute workloads among specialized agents (e.g., managers, executors, reviewers), merging their outputs for robust planning. (4) Learning-based Enhancement: distinct from the prompt-engineering approaches above, recent works have also explored reinforcement learning to directly optimize the planning capabilities of models through trial-and-error~\cite{ning2025deeptravel,zhang2025agent}.

\section{Primitive Tasks}

\subsection{Problem Formulation}

To rigorously formalize the planning process, we define the variable spaces involved. Let $\mathcal{Q}$ denote the space of natural language user queries, and $\mathcal{C}$ be the space of derived constraints. To model the interaction with the external environment, we define $\mathcal{T}$ as the space of tool invocations. The sandbox function $S(\cdot)$ takes an action from $\mathcal{T}$ as input to retrieve content from the space of external information $\mathcal{I}$ (i.e., $S: \mathcal{T} \to \mathcal{I}$). Finally, let $\mathcal{P}$ represent the space of generated plans (itineraries), and $\mathcal{E}$ denote the space of diagnostic error feedback.

The five sub-tasks are formalized as follows:

\textbf{Constraint Extraction.} At the initial stage, the LLM is required to extract constraints from the natural language query.
\begin{equation*}
    \text{Constraint Extraction}: \mathcal{Q} \to \mathcal{C},
\end{equation*}
where for any $q \in \mathcal{Q}$, the output is a constraint set $C_q \subseteq \mathcal{C}$. Following the setup in ChinaTravel~\cite{shao2026chinatravel}, we convert the constraints in each benchmark into a unified Domain Specific Language (DSL) representation to enable consistent evaluation. Constraint extraction reflects a model’s understanding of the core task requirements and can influence a series of subsequent actions of the language model, such as discarding queries that are deemed uninformative or systematically verifying the validity of a plan during plan generation and self-checking.

\textbf{Tool Use.} To bridge the gap between internal knowledge and real-time world states (which are typically simulated in benchmarks), the agent must actively retrieve external information~\cite{qintoolllm,shen2024llm}.
\begin{equation*}
    \text{Tool Use}: \mathcal{Q} \to \mathcal{T} \xrightarrow{S} \mathcal{I},
\end{equation*} where the agent first generates tool invocations $T_q \in \mathcal{T}$ based on query $q$, which are then executed by the Sandbox $S$ to yield the information context $I_q \subseteq \mathcal{I}$. This step is critical for real-world planning, as it determines the boundary of information available to the planner. A failure here inevitably leads to hallucinations or plans based on outdated assumptions (e.g., non-existent flights).

\textbf{Plan Generation.} Conditioned on the user's intent and the retrieved context, the agent synthesizes a complete itinerary.
\begin{equation*}
    \text{Plan Generation}: \mathcal{Q} \times \mathcal{C} \times \mathcal{I} \to \mathcal{P},
\end{equation*} where the model produces a candidate plan $p \in \mathcal{P}$ that aims to satisfy the constraint set $C_q$ utilizing information $I_q$. This process represents the core reasoning engine of the agent, primarily requiring it to synthesize a feasible itinerary that strictly satisfies the extracted constraint set $C_q$.

\textbf{Error Identification.} Error identification module serves as a verifier to diagnose the feasibility and logical soundness of the generated plan.
\begin{equation*}
    \text{Error Identification}: \mathcal{Q} \times \mathcal{C} \times \mathcal{I} \times \mathcal{P} \to \mathcal{E},
\end{equation*} where the input tuple is analyzed to output a specific error feedback set $E_p \subseteq \mathcal{E}$. Unlike binary success detection, this module explicitly isolates reasoning deficits by classifying violations into specific categories (e.g., repeated activities, violations of user requirements, or hallucinations), thereby providing fine-grained feedback for self-correction.

\textbf{Error Correction.} Finally, the agent acts on the diagnostic feedback to repair the erroneous itinerary without human intervention.
\begin{equation*}
    \text{Error Correction}: \mathcal{Q} \times \mathcal{C} \times \mathcal{I} \times \mathcal{P} \times \mathcal{E} \to \mathcal{P},
\end{equation*} where the agent transforms the initial faulty plan $p$ and error feedback $E_p$ into a refined plan $p' \in \mathcal{P}$. This capability tests the model's robustness and its ability to incorporate negative feedback to adjust its reasoning path, preventing the persistence of errors in the final output.

\subsection{Data Construction}

For data construction, we select three benchmarks with available annotated data: TravelPlanner~\cite{xie2024travelplanner}, ChinaTravel~\cite{shao2026chinatravel}, and TripCraft~\cite{chaudhuri2025tripcraft}, to facilitate the subsequent construction of oracle intermediate contexts. We use the validation split of TravelPlanner and the human subset of ChinaTravel, and sample a subset of TripCraft.

For the \textbf{constraint extraction} stage, we convert the template-based constraints in TravelPlanner and TripCraft into the corresponding DSL representations, while directly adopting the annotated constraints provided by ChinaTravel. For \textbf{tool use}, we provide the LLM with the corresponding tool documentation and require it to further refine the query to specify concrete tool use requirements, while generating the corresponding tool calls, whose syntactic correctness is then verified in a sandbox environment. The prompt used for tool call generation is provided in~\cref{app:tool_use_generation_prompt}. 

For \textbf{plan generation}, we adopt the methodologies from TTG~\cite{ju-etal-2024-globe}, utilizing an MILP solver for TravelPlanner and TripCraft, and the UrbanTrip~\cite{chinatravel2025} strategy for ChinaTravel.
We formulate the solver's task as a feasibility problem. The solver seeks a plan $p$ within the feasible region defined by the constraint set $C_q$:
\begin{equation*}
\begin{aligned}
\text{find} \quad & p \in \mathcal{P} \\
\text{s.t.} \quad & \forall c \in C_q, \ p \models c, 
\end{aligned}
\end{equation*}
where the notation $p \models c$ indicates that the generated plan $p$ strictly satisfies the constraint $c$. We construct the information context $I_q$ by augmenting a minimal sufficient set $I_{q, \text{min}}$, comprised of all PoI entities in the reference plan, with varying levels of distractor information $I_{q, \text{dist}}$. Specifically, $I_{q, \text{dist}}$ represents valid candidate PoI entities identified by the solver. In the minimal setting, $I_{q, \text{dist}}$ remains empty; whereas in the moderate and rich settings, we incorporate 10 and 20 additional PoIs per category (i.e. attraction, accommodation and restaurant) for each visited city, respectively.

Similarly, for \textbf{error identification} and \textbf{error correction}, to construct a plan containing specific errors, we define a target violation set $C_{q, neg} \subseteq C_p$. The solver seeks a plan $p_f$ that strictly violates these selected constraints:
\begin{equation*}
\begin{aligned}
\text{find} \quad & p_f \in \mathcal{P} \\
\text{s.t.} \quad & \forall c \in C_{q, neg}, \ p_f \models \neg c \\
& \forall c \in C \setminus C_{q, neg}, \ p_f \models c,
\end{aligned}
\end{equation*} where the notation $p_f \models \neg c$ indicates that the generated plan $p_f$ strictly entails the negation of the constraint $c$ (i.e., violates $c$). During the diagnostic phase, error identification takes the minimal input $I_{q, \text{min}}$, and error correction uses an augmented input $I_q' = I_{q, \text{min}} \cup I_{q, \text{oracle}}$, incorporating necessary oracle details from the ground truth plan.

\subsection{Evaluation Metrics}

For constraint extraction and error identification, we evaluate Precision, Recall, F1 score, and Exact Match. For tool use, we measure tool name accuracy, argument accuracy, and overall accuracy. For plan generation and correction, we adopt the plan evaluation metrics provided by the corresponding benchmarks.

\section{Results}

We evaluate a set of models, including GPT-5.2, DeepSeek V3.2 (DS V3.2), Qwen3 Max, Claude Sonnet 4.5 (Claude S4.5), and Gemini 3 Pro Preview (Gemini 3 Pro), in both reasoning and non-reasoning modes. For constraint extraction and plan generation, our experiments focus on non-reasoning or low-reasoning models. Due to API stability issues, Gemini 3 Pro Preview High was not evaluated on plan generation for TripCraft. 
Furthermore, in ChinaTravel, including only the inner-city transportation information in fact tells the model how to select and arrange POIs, essentially completing the planning. The potential number of inner-city transportation entries scales roughly with the square of the number of POIs. Even after removing obviously impossible connections (e.g., between accommodations), the token count of the minimal information context $I_{q, \text{min}}$ still exceeds 64k. This makes testing the results extremely difficult, and therefore ChinaTravel was not evaluated for plan generation and error correction tasks.

\subsection{Constraint Extraction}

\begin{table*}[t]
\centering
\small
\caption{LLM constraint extraction performance. We report precision (P), recall (R), f1 score (F1), and exact match (EM). The results show that LLMs perform well in simpler scenarios, but their performance noticeably drops as the scenarios become more complex.}
\begin{tabular}{lcccccccccccc}
\hline

& \multicolumn{4}{c}{\cellcolor{mygreen!75}TravelPlanner}
& \multicolumn{4}{c}{\cellcolor{mygreen!75}TripCraft}
& \multicolumn{4}{c}{\cellcolor{mygreen!75}ChinaTravel} \\
Model
& P & R & F1 & EM
& P & R & F1 & EM
& P & R & F1 & EM \\
\hline
GPT-5.2
& 0.95 & 0.94 & 0.95 & 0.76
& 0.90 & 0.93 & 0.92 & 0.62
& 0.63 & 0.58 & 0.61 & 0.00 \\
DS V3.2
& 0.96 & 0.96 & 0.96 & 0.85
& 0.89 & 0.92 & 0.91 & 0.57
& 0.62 & 0.63 & 0.63 & 0.00 \\
Qwen3 Max
& 0.96 & 0.94 & 0.95 & 0.79
& 0.92 & 0.94 & 0.93 & 0.68
& \textbf{0.74} & 0.56 & 0.64 & 0.00 \\
Claude S4.5
& 0.98 & 0.97 & 0.97 & 0.86
& \textbf{0.94} & \textbf{0.97} & \textbf{0.96} & \textbf{0.73}
& 0.72 & 0.65 & 0.69 & 0.00 \\
Gemini 3 Pro
& \textbf{0.98} & \textbf{0.97} & \textbf{0.98} & \textbf{0.88}
& 0.89 & 0.91 & 0.90 & 0.55
& 0.65 & \textbf{0.78} & \textbf{0.71} & 0.00 \\
\hline
\end{tabular}
\label{tab:constraint_extraction_res}
\end{table*}

\begin{figure}
    \centering
    \includegraphics[width=\linewidth]{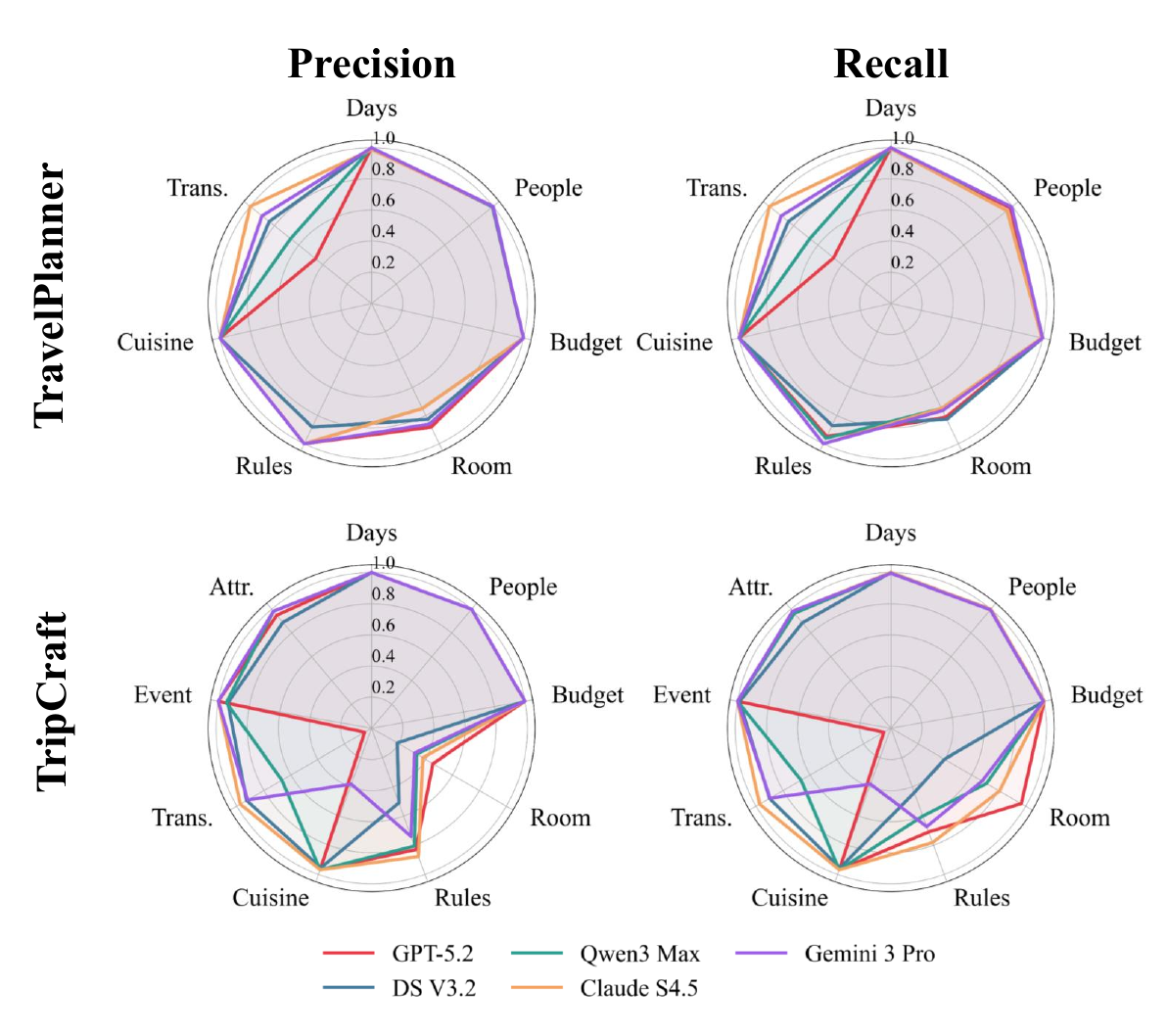}
    \caption{Constraint-wise precision and recall on TravelPlanner and TripCraft. The results show that LLMs have relatively poor ability to extract constraints expressed in non-numerical terms.}
    \label{fig:constraint_pr}
\end{figure}

The constraint extraction results are presented in \cref{tab:constraint_extraction_res}, where GPT-5.2 was employed to verify whether the extracted DSL aligns with the annotated constraints. The prompt for the LLM as a judge is provided in~\cref{app:llm_as_a_judge_prompt}. This evaluation reveals a stark performance dichotomy. On the TravelPlanner benchmark, state-of-the-art models demonstrate near-saturated performance, with Gemini 3 Pro achieving an F1 score of 0.98 and an Exact Match (EM) rate of 0.88. This indicates that current LLMs have effectively mastered constraint extraction in simplified scenarios.

\textbf{Imperfect extraction due to complexity.} As the evaluation transitions to TripCraft, the limits of model capability become evident. Despite maintaining high F1 scores (peaking at 0.96 with Claude Sonnet 4.5), the Exact Match rates exhibit a noticeable decline, ranging between 0.55 and 0.73. This divergence—robust F1 but deteriorating EM—signals that as scenario complexity increases, models struggle to flawlessly capture the complete constraint set, even if they successfully retrieve the majority of individual requirements.

\textbf{Open-constraint and expression diversity challenges.} The performance landscape shifts drastically on the ChinaTravel dataset, marking a precipitous drop. F1 scores fall to the 0.61--0.71 range, but most critically, the dataset yields a universal 0.00 EM rate across all evaluated models. Unlike the TravelPlanner, where models could frequently align perfectly with user intents, this absolute failure in exact matching implies an insurmountable complexity barrier in open-world scenarios. It suggests that while models can recover partial information (reflected in moderate Recall), they systematically fail to reconstruct the precise, holistic boundary of intricate user needs. Further analysis in \cref{fig:constraint_pr} shows that for constraints with strict numerical values, such as days, people\_number, and budget, models generally extract them accurately with few or no omissions. In contrast, extraction performance degrades for constraints whose values are text or lists of text, especially those involving negation (room\_type, house\_rule and transportation).

Ultimately, these findings expose a critical deficiency in the generalization capabilities of existing models when confronting open-ended constraints. The inability to maintain performance across varying levels of complexity suggests that current models heavily rely on the explicit coverage of constraint types provided in the prompt, rather than robustly identifying constraints beyond those specifications.

\subsection{Tool Use}

\begin{table*}[t]
\small
\centering
\caption{LLM tool use performance. 
We report accuracy (overall), parameter accuracy (correctness of tool arguments), and tool accuracy (correct tool selection). The results show that LLMs can invoke tools almost perfectly in simple scenarios, but they sometimes make errors in complex scenarios. }

\begin{tabular}{lcccccccccccc}
\hline
& \multicolumn{3}{c}{\cellcolor{myyellow!75}TravelPlanner}
& \multicolumn{3}{c}{\cellcolor{myyellow!75}TripCraft}
& \multicolumn{3}{c}{\cellcolor{myyellow!75}ChinaTravel} \\
Model & Acc & Param Acc & Tool Acc & Acc & Param Acc & Tool Acc & Acc & Param Acc & Tool Acc \\
\hline
\multicolumn{10}{l}{\textit{Non-Reasoning}} \\
GPT-5.2 & 1.00 & 1.00 & 1.00 & 1.00 & 1.00 & 1.00 & \textbf{0.84} & \textbf{0.90} & \textbf{0.84} \\
Claude S4.5 & 1.00 & 1.00 & 1.00 & 1.00 & 1.00 & 1.00 & 0.71 & 0.75 & 0.71 \\
DS V3.2 & 1.00 & 1.00 & 1.00 & 1.00 & 1.00 & 1.00 & 0.70 & 0.72 & 0.70 \\
Qwen3 Max & 1.00 & 1.00 & 1.00 & 1.00 & 1.00 & 1.00 & 0.65 & 0.67 & 0.65 \\

\hline
\multicolumn{10}{l}{\textit{Reasoning Mode}} \\
GPT-5.2 & 1.00 &1.00 & 1.00 & 1.00 & 1.00 & 1.00 & \textbf{0.84} & \textbf{0.91} & \textbf{0.84} \\
Claude S4.5 & 1.00 & 1.00 & 1.00 & 1.00 & 1.00 & 1.00 & 0.71 & 0.75 & 0.71 \\
DS V3.2 & 1.00 & 1.00 & 1.00 & 1.00 & 1.00 & 1.00 & 0.74 & 0.82 & 0.74 \\
Qwen3 Max & 1.00 & 1.00 & 1.00 & 1.00 & 1.00 & 1.00 & 0.64 & 0.66 & 0.64 \\
Gemini 3 Pro Low    & 1.00 & 1.00 & 1.00 & 1.00 & 1.00 & 1.00 & 0.82 & 0.90 & 0.82 \\
Gemini 3 Pro High   & 1.00 & 1.00 & 1.00 & 1.00 & 1.00 & 1.00 & 0.82 & 0.90 & 0.82 \\
\hline
\end{tabular}
\label{tab:tool_use}
\end{table*}

\begin{figure}
    \centering
    \includegraphics[width=\linewidth]{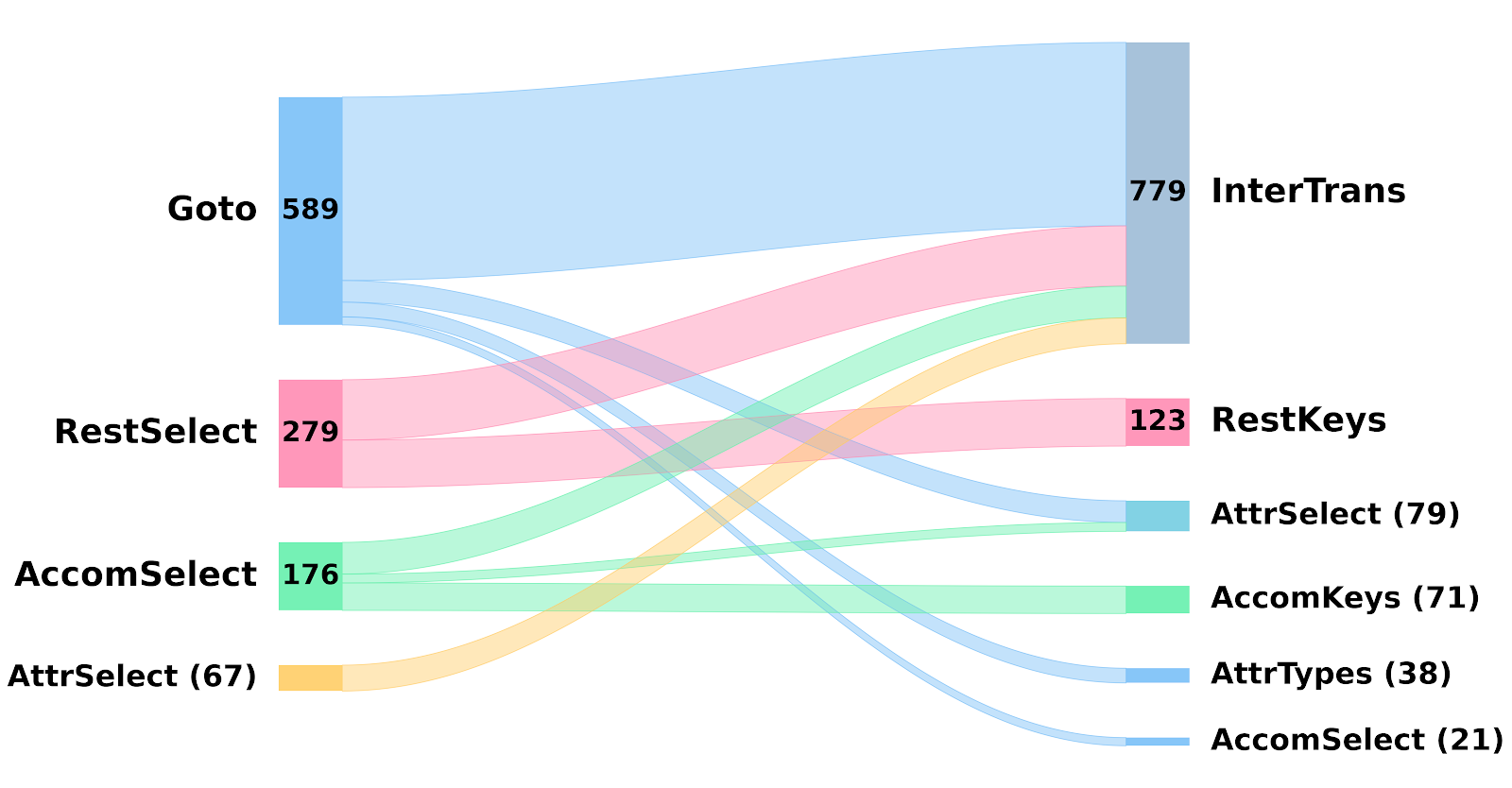}
    \caption{Top 10 specific confusion pairs (oracle $\rightarrow$ predicted). The results show that LLMs are prone to confusing similar tools in complex scenarios.}
    \label{fig:cnts_confusion}
\end{figure}

The complete results of tool use are shown in~\cref{tab:tool_use}. Tool use in existing travel planning benchmarks is relatively decoupled from planning itself. Most benchmarks adopt a retrieve-then-plan paradigm, where tools mainly serve as information providers and do not require substantial reasoning during invocation, leading to saturated performance on TravelPlanner and TripCraft across all models. In contrast, on ChinaTravel, the increased number of tools, the presence of tools with similar interfaces but distinct functionalities, and the more complex planning scenarios lead to occasional tool use errors.

\textbf{Confusion between similar tools.} The top ten most frequent tool misuses are shown in~\cref{fig:cnts_confusion}. Given that the increase in context length is marginal (approximately 600 tokens) compared to other benchmarks where the model achieves 100\% accuracy, it is unlikely that context overload is the primary cause of the observed confusion. Instead, the misclassification of \texttt{Goto} as \texttt{IntercityTransport} indicates that the model struggles to clearly distinguish between innercity and intercity transportation. Furthermore, regarding the misclassification of other non-transport tools into \texttt{IntercityTransport}, we attribute this to a tendency towards over-planning. In complex scenarios, the model likely misinterprets atomic tool queries as implicit requests for a holistic travel plan. Consequently, it treats \texttt{IntercityTransport} as the logical entry point to initiate the imagined itinerary.

\subsection{Plan Generation}

\begin{table*}[t]
\small
\centering
\caption{LLM plan generation performance across different settings. The results indicate that there is still significant room for improvement for LLMs in plan generation tasks.}
\label{tab:benchmark_results}

\begin{tabular*}{0.9\textwidth}{@{\extracolsep{\fill}}lcccccc}
\toprule
\multirow{2}{*}{\textbf{Model}} & \multicolumn{3}{c}{\textbf{\cellcolor{myred!75}TravelPlanner}} & \multicolumn{3}{c}{\textbf{\cellcolor{myred!75}TripCraft}} \\
\cmidrule(lr){2-4} \cmidrule(lr){5-7}
 & \textbf{Minimal} & \textbf{Moderate} & \textbf{Rich} & \textbf{Minimal} & \textbf{Moderate} & \textbf{Rich} \\
\midrule
\textbf{GPT-5.2} & 
0.13 & 0.18 & 0.20 & 
0.03 & 0.06 & 0.01 \\
\textbf{Claude S4.5} & 
0.23 & 0.26 & 0.21 & 
0.03 & 0.07 & 0.04 \\
\textbf{DS v3.2} & 
0.27 & 0.24 & 0.25 & 
0.05 & \textbf{0.13} & \textbf{0.15} \\
\textbf{Qwen3 Max} & 
0.21 & 0.17 & 0.19 & 
0.01 & 0.03 & 0.04 \\
\textbf{Gemini 3 Pro Low} & 
0.32 & \textbf{0.42} & \textbf{0.43} & 
\textbf{0.06} & 0.07 & 0.02 \\

\textbf{Gemini 3 Pro High} & 
\textbf{0.34} & \textbf{0.42} & 0.35 & 
-- & -- & -- \\

\bottomrule
\end{tabular*}%
\end{table*}

Plan generation is a core capability in travel planning, where models are required to organize and combine extracted constraints with collected information to produce a feasible plan. During this process, the model must perform substantial reasoning to ensure that each POI appears in an appropriate position, while the combined effects, such as total cost, remain within the specified budget.

\cref{tab:benchmark_results} shows the final plan success rates under different amounts of provided valid information and detailed error information is provided in~\cref{app:detail_plan_generation_res}. It is evident that there remains substantial room for improvement in models’ ability to generate valid plans. Even the best-performing model, Gemini 3 Pro, achieves a success rate of only 43\% on TravelPlanner. In TripCraft, models are additionally required to generate a visitation schedule for each POI. Although there are no strict travel-time constraints, the results indicate that simply producing an ordered list of visits along with the nearby transportation hubs already poses a significant challenge for the models. 


\begin{figure}[tb]
    \centering
    \includegraphics[width=0.9\linewidth]{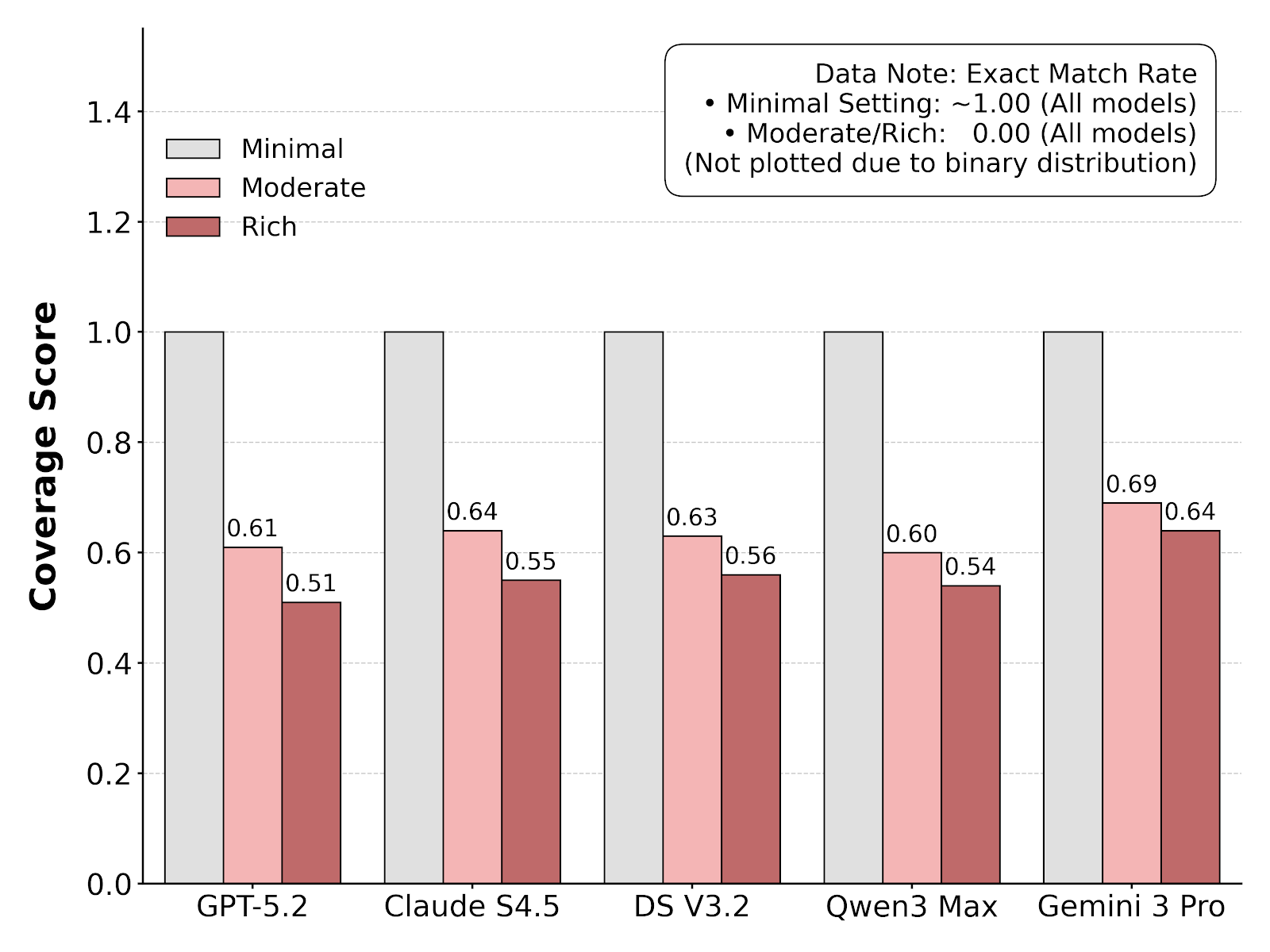}
    \caption{POI match rate and coverage across different settings. The results show that LLMs tend to favor certain POIs over others when generating plans.}
    \label{fig:poi_cover}
\end{figure}

\textbf{Performance sensitive to POI selection.} However, surprisingly, overall model performance under the minimal setting, with only the valid information extracted from the reference plan, is worse than under the moderate setting, where additional information not present in the reference answers is provided. We present the overlap between the POIs in the model-generated correct plans and those in the reference plans in \cref{fig:poi_cover}. The Match Rate calculates the proportion of fully matched POIs in the entire plan relative to the total number of correct plans. The Coverage computes the average ratio of overlapping POIs to the total number of POIs in the reference plans. It is obvious that the plans generated by the model often differ significantly from the reference plans. This suggests that the model's POI selection seems to be influenced by some prior knowledge. When there are alternative choices available, the overlap of POIs between the model's generated plan and the reference plan decreases significantly. We also selected 37 queries where Gemini 3 Pro failed in the minimal setting but succeeded in other settings, with POI coverage below 0.6. By reconstructing the minimal-level input based on the model-generated plans, we found that the success rate of Gemini 3 Pro on these 37 queries increased from 0 to 0.49. This also indicates that selecting specific POIs can significantly improve the model's performance.

\subsection{Error Identification}

\begin{table*}[t]
\small
\centering
\caption{LLM error identification performance (Ex indicates that there are x errors in the plan). The results show that LLMs can identify errors in plans, but they often detect errors that do not actually exist.}
\label{tab:err_id}
\setlength{\tabcolsep}{5pt}
\begin{tabular}{lccccc}
\toprule
\textbf{Model} & \textbf{\cellcolor{myorange!75}ChinaTravel} & \textbf{\cellcolor{myorange!75}TripCraft} & \textbf{\cellcolor{myorange!75}TP (E1)} & \textbf{\cellcolor{myorange!75}TP (E2)} & \textbf{\cellcolor{myorange!75}TP (E3)} \\
\cmidrule(lr){2-6}
& \multicolumn{5}{c}{\small \textit{Metrics: Precision / Recall / F1}} \\
\midrule
\multicolumn{6}{l}{\textit{Non-Reasoning}} \\
GPT-5.2      & 0.65 / \textbf{0.93} / 0.77 & 0.62 / 0.80 / 0.70 & 0.64 / \textbf{0.85} / 0.73 & 0.79 / \textbf{0.77} / 0.78 & 0.84 / \textbf{0.78} / 0.81 \\
Claude S4.5  & \textbf{0.69} / 0.92 / \textbf{0.79} & 0.79 / 0.73 / 0.76 & 0.79 / 0.78 / 0.78 & \textbf{0.94} / 0.73 / \textbf{0.82} & \textbf{0.95} / 0.75 / \textbf{0.84} \\
DS V3.2      & 0.66 / 0.90 / 0.76 & 0.73 / \textbf{0.82} / 0.77 & 0.75 / 0.77 / 0.76 & 0.85 / \textbf{0.77} / 0.81 & 0.91 / \textbf{0.78} / \textbf{0.84} \\
Qwen3 Max    & 0.58 / 0.92 / 0.71 & \textbf{0.83} / 0.77 / \textbf{0.80} & \textbf{0.86} / 0.78 / \textbf{0.82} & \textbf{0.94} / 0.73 / \textbf{0.82} & 0.94 / 0.73 / 0.82 \\
\midrule
\multicolumn{6}{l}{\textit{Reasoning Mode}} \\
GPT-5.2       & 0.69 / \textbf{0.97} / \textbf{0.81} & 0.65 / 0.83 / 0.73 & 0.73 / \textbf{0.88} / 0.80 & 0.89 / \textbf{0.81} / \textbf{0.85} & 0.91 / \textbf{0.82} / \textbf{0.86} \\
Claude S4.5   & 0.68 / 0.91 / 0.78 & 0.80 / 0.73 / 0.76 & 0.79 / 0.78 / 0.78 & \textbf{0.95} / 0.72 / 0.82 & \textbf{0.95} / 0.75 / 0.83 \\
DS V3.2       & \textbf{0.70} / 0.93 / 0.80 & 0.75 / 0.83 / 0.79 & 0.72 / 0.80 / 0.76 & 0.87 / 0.77 / 0.81 & 0.91 / 0.77 / 0.83 \\
Qwen3 Max     & 0.44 / 0.95 / 0.60 & \textbf{0.84} / 0.76 / \textbf{0.80} & \textbf{0.80} / 0.81 / \textbf{0.81} & 0.86 / 0.74 / 0.79 & \textbf{0.95} / 0.73 / 0.83 \\
Gemini 3 Pro Low  & 0.66 / 0.89 / 0.76 & 0.73 / 0.82 / 0.77 & 0.75 / 0.85 / 0.79 & 0.89 / 0.77 / 0.83 & 0.92 / 0.78 / 0.84 \\
Gemini 3 Pro High & 0.67 / 0.88 / 0.76 & 0.75 / \textbf{0.84} / 0.79 & 0.74 / 0.84 / 0.79 & 0.89 / 0.73 / 0.80 & 0.92 / 0.79 / 0.85 \\
\bottomrule
\end{tabular}
\end{table*}

If the model fails to generate the correct result in one go, identifying errors becomes especially important in the process of self-correction. Only by accurately identifying errors can precise corrections be made in subsequent processes. In \cref{tab:err_id}, we present the model's performance in identifying errors within the plans without being explicitly informed of the number of errors. Here, TravelPlanner (E2) and TravelPlanner (E3) indicate the presence of 2 and 3 errors, respectively, while the rest have one error each. 

\textbf{Overestimation of Errors.} The consistently high recall in the results indicates that the model is generally able to identify the errors. However, by observing the performance on TravelPlanner, it can be seen that as the number of errors increases, the precision of the identifications gradually improves. This suggests that the LLM is capable of recognizing errors in the plan but tends to believe that there is more than one error in the plan. This tendency is also evident in certain specific examples. The model easily interprets \texttt{occupancy} as a type of \texttt{house\_rule}, assuming that the rule is violated when the number of travelers exceeds the maximum capacity of a single room, rather than recognizing that the solution could involve booking additional rooms. Similarly, for \texttt{room\_type}, the model may mistakenly assumes that a \texttt{private room} does not meet the requirement of \texttt{not shared room}.

\subsection{Error Correction}

\begin{table*}[t]
\small
\centering
\caption{LLM error correction performance. The results show that LLMs face certain difficulties in error correction.}
\label{tab:err_correction_res}
\setlength{\tabcolsep}{5pt}
\begin{tabular}{lccccccccc}
\toprule
\multirow{2}{*}{\textbf{Model}} & \textbf{\cellcolor{mypurple!75}TripCraft} & \textbf{\cellcolor{mypurple!75}TP (E1)} & \textbf{\cellcolor{mypurple!75}TP (E2)} & \textbf{\cellcolor{mypurple!75}TP (E3)} & & \textbf{\cellcolor{mypurple!75}TripCraft} & \textbf{\cellcolor{mypurple!75}TP (E1)} & \textbf{\cellcolor{mypurple!75}TP (E2)} & \textbf{\cellcolor{mypurple!75}TP (E3)} \\

\cmidrule(lr){2-5} \cmidrule(lr){7-10}
 & \multicolumn{4}{c}{\textit{Non-Reasoning}} & & \multicolumn{4}{c}{\textit{Reasoning Mode}} \\
\midrule
GPT-5.2      & 0.05 & 0.06 & 0.05 & 0.03 & & 0.05 & 0.10 & 0.09 & 0.18 \\
Claude S4.5  & 0.07 & 0.13 & 0.08 & 0.08 & & 0.06 & 0.15 & 0.09 & 0.08 \\
DS V3.2     & \textbf{0.14} & \textbf{0.22} & \textbf{0.21} & \textbf{0.17} & & 0.06 & 0.14 & 0.21 & 0.21 \\
Qwen3 Max    & 0.07 & 0.17 & 0.11 & 0.05 & & 0.04 & 0.16 & 0.10 & 0.08 \\
Gemini 3 Pro Low     & - & - & - & - & & 0.11 & \textbf{0.25} & 0.25 & \textbf{0.32} \\
Gemini 3 Pro High     & - & - & - & - & & \textbf{0.12} & 0.24 & \textbf{0.26} & 0.31 \\
\bottomrule
\end{tabular}
\end{table*}

\begin{figure}
    \centering
    \includegraphics[width=0.85\linewidth]{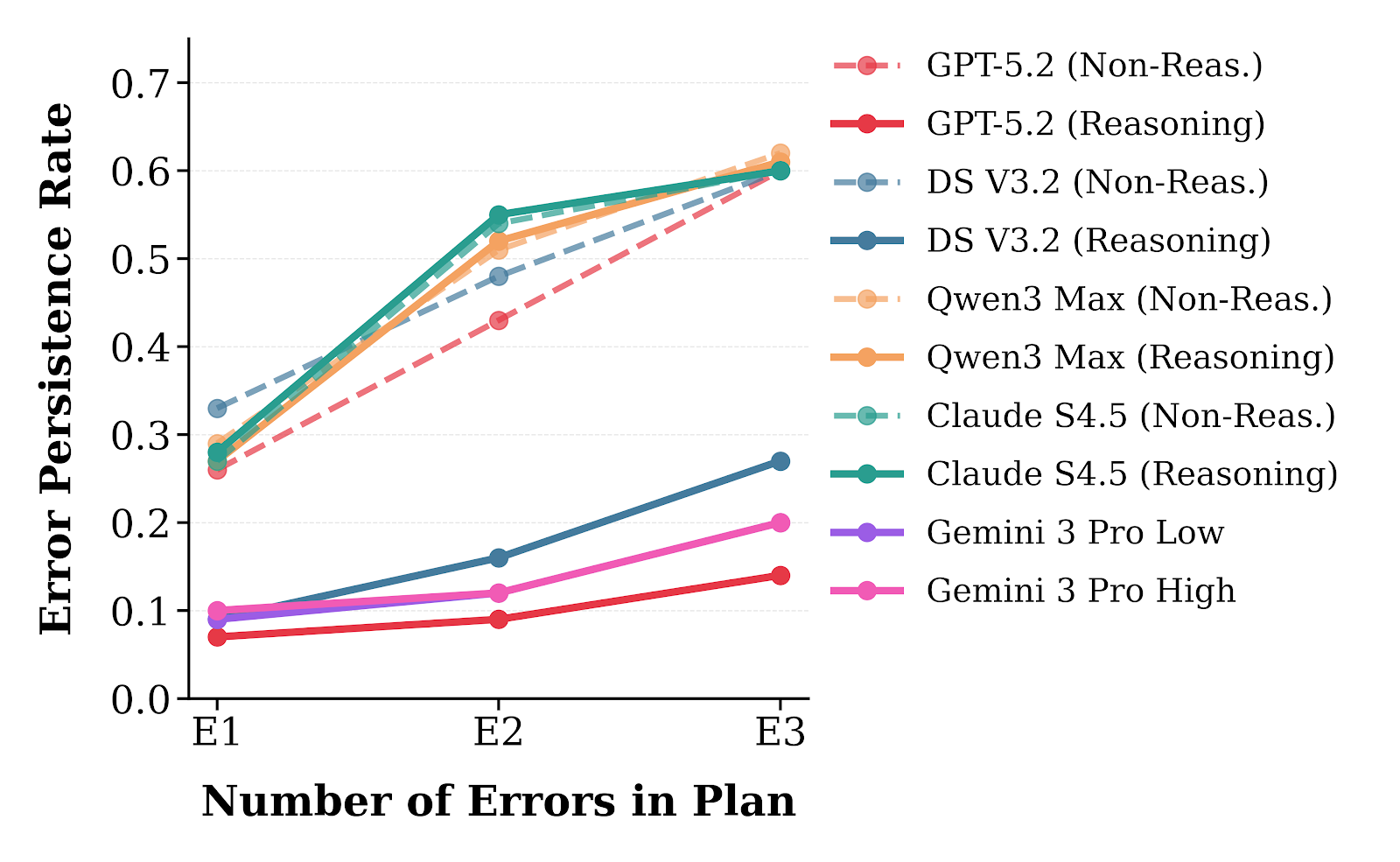}
    \caption{Error persistence on TravelPlanner. The results show that non-reasoning models and some reasoning models struggle to avoid existing errors.}
    \label{fig:err_persistence}
\end{figure}

Error correction requires the model to locate the positions of the errors and modify them accordingly. Some errors may be caused by combinations of multiple POIs, and modifying any single POI will not resolve the issue. \cref{tab:err_correction_res} shows the results of error correction, which are generally comparable to the plan generation results under the easy setting.

\textbf{Error Persistence.} Observing the plans before and after correction, we find a certain degree of error persistence, meaning that despite pointing out the errors in the provided plan, the same errors still appear in the model's corrected plan. \cref{fig:err_persistence} demonstrates this phenomenon which is particularly pronounced in non-reasoning mode. Furthermore, as the number of errors increases, it becomes progressively more difficult to completely avoid all past errors. Interestingly, we observe that GPT-5.2 consistently maintains the lowest error persistence rate. However, its corrected plans perform poorly, suggesting that the model may be overly focused on avoiding the same mistakes, to the extent that it somewhat neglects the overall validity of the plan.

\section{Limitations and Future Work}

Despite the comprehensive analysis presented, our study has several limitations.
First, we adopt an atomic evaluation strategy with oracle inputs to isolate individual reasoning modules. While this design enables fine grained diagnosis and avoids error propagation, it does not capture error accumulation or iterative correction across modules.
Second, the evaluation scale is limited by the substantial cost of large scale evaluation. Moreover, although prior work has explored training based improvements for specific aspects of the task, a systematic investigation of whether fine tuning or reinforcement learning can bridge the full range of observed capability gaps remains lacking \cite{zhang2025agent,fang2025memp}.
Finally, our framework considers a static text based environment, whereas real world planning often involves multimodal inputs and dynamic updates.

Future research may further investigate the evaluation of LLM based planning under more dynamic settings with compositional subtask structures. Evaluating how models coordinate multiple interdependent subtasks, adapt to intermediate state changes, and manage partial information could provide deeper insights into error accumulation, correction behaviors, and long horizon planning consistency.
In parallel, improving the planning performance of LLMs and LLM based agents remains an important direction. Limitations in long context understanding, memory, and state tracking can substantially increase task difficulty in complex travel planning scenarios such as ChinaTravel. Future work could explore reinforcement learning with richer environments and finer-grained reward mechanisms, architectural improvements, enhanced long-context modeling, and explicit memory mechanisms to help models better satisfy complex and interacting constraints.



\section{Conclusion}

In this work, we presented a systematic framework to dissect the long-horizon reasoning capabilities of LLMs by decomposing travel planning into atomic sub-tasks. Through this decoupled evaluation framework, we meticulously analyzed specific deficits across atomic sub-capabilities, offering valuable insights to inform future model enhancements.

Our empirical analysis reveals a nuanced capability profile. While models demonstrate competence in basic tool usage and explicit instruction following, they falter in complex scenarios. Some issues also arise in the downstream stages, where plan generation is hampered by structural biases and sensitivity to POI selection. Furthermore, the self-correction mechanism exhibits some limitations: models exhibit excessive sensitivity during error identification, often flagging issues that do not exist, while simultaneously failing to effectively rectify actual mistakes or avoid error repetition in revised plans. Overall, these findings underscore that the reasoning capabilities of LLMs, particularly under complex constraints, still require significant improvement.

\section*{Impact Statement}

This paper presents a diagnostic study on the planning capabilities of LLMs. The goal is to identify current limitations in reasoning and reliability, thereby fostering the development of more capable LLMs for real-world applications.




\bibliography{example_paper}
\bibliographystyle{icml2026}

\newpage
\appendix
\onecolumn

\section{Statistics of Errors in Plan Generation}
\label{app:detail_plan_generation_res}
\begin{figure}[htbp]
    \centering
    \includegraphics[width=0.77\linewidth]{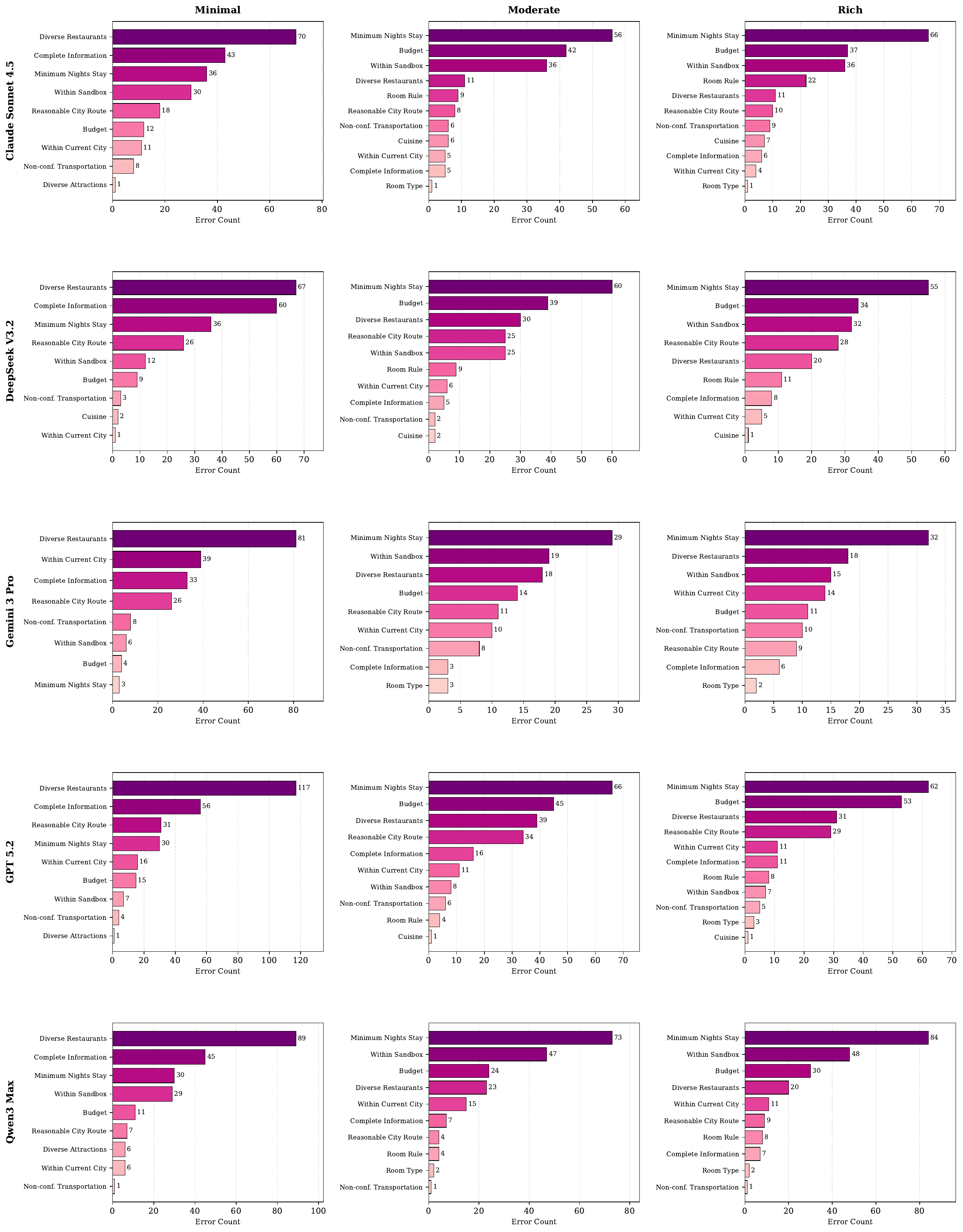}
    \caption{Statistics of plan error types under different settings on TravelPlanner}
    \label{fig:plan_generation_err_tp}
\end{figure}

\begin{figure}[htbp]
    \centering
    \includegraphics[width=0.77\linewidth]{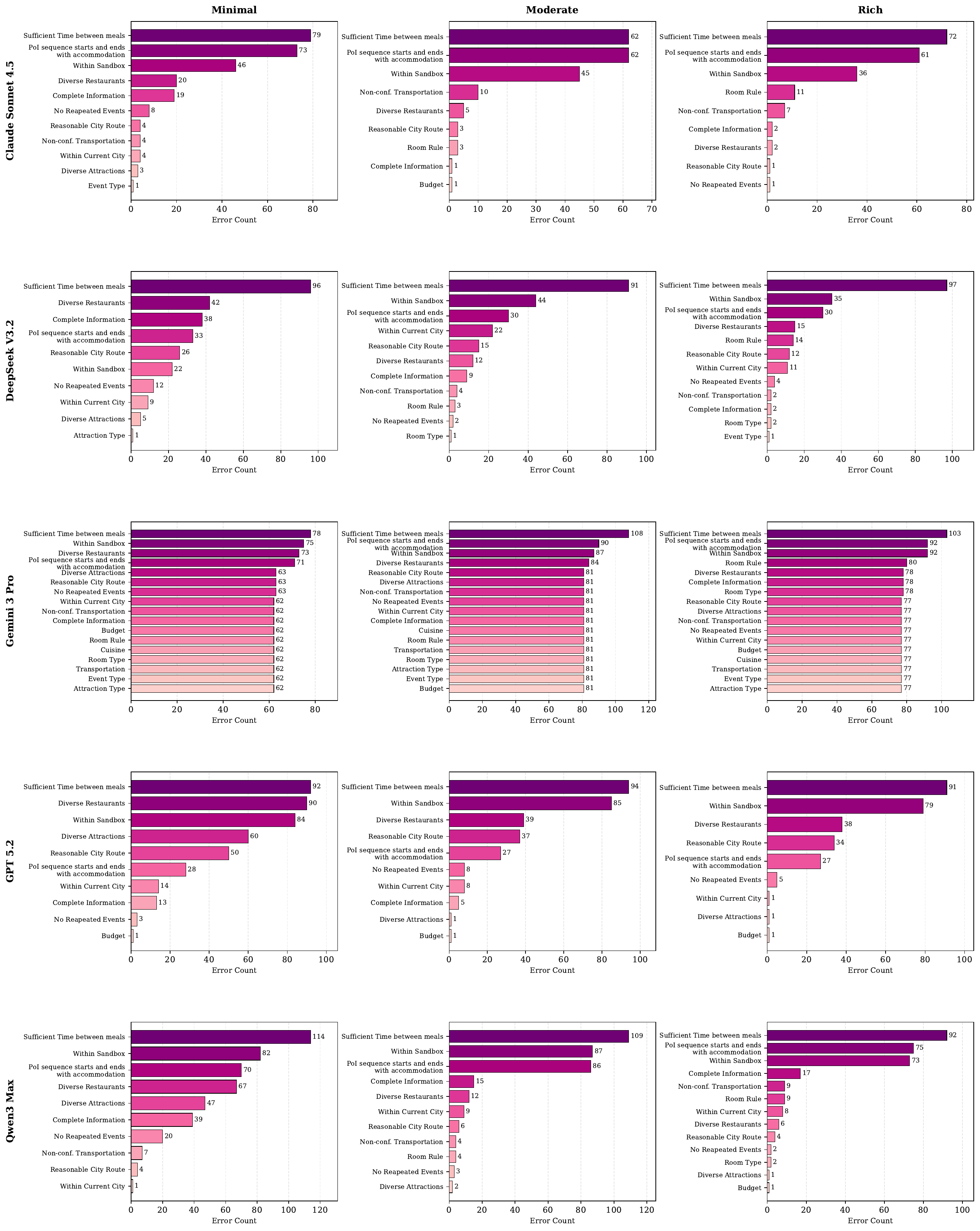}
    \caption{Statistics of plan error types under different settings on TripCraft}
    \label{fig:plan_generation_err_tc}
\end{figure}

\cref{fig:plan_generation_err_tp} and \cref{fig:plan_generation_err_tc} show the specific causes of errors in the plans generated by the model. On TripCraft, Gemini 3 Pro exhibits a high number of all error types due to the large number of solutions with formatting errors.

\FloatBarrier

\section{Exact Match Accuracy of Error Identification}

\begin{table}[htbp]
\centering
\caption{Error Identification Performance (Exact Match) }
\label{tab:exact_match_acc_decimal}
\setlength{\tabcolsep}{5pt}
\begin{tabular}{lccccc}
\toprule
\textbf{Model} & \textbf{ChinaTravel} & \textbf{TripCraft} & \textbf{TP (E1)} & \textbf{TP (E2)} & \textbf{TP (E3)} \\
\cmidrule(lr){2-6}
& \multicolumn{5}{c}{\small \textit{Metric: Exact Match Accuracy}} \\
\midrule
\multicolumn{6}{l}{\textit{Non-Reasoning}} \\
GPT-5.2            & 0.69 & 0.47 & 0.51 & 0.40 & 0.25 \\
GPT-5.2 Medium     & 0.70 & 0.56 & 0.61 & \textbf{0.49} & \textbf{0.36} \\
Claude S4.5        & 0.71 & 0.62 & 0.67 & 0.46 & 0.35 \\
DS V3 (Chat)       & 0.66 & 0.59 & 0.58 & 0.47 & \textbf{0.36} \\
Qwen3 Max          & 0.74 & \textbf{0.67} & \textbf{0.70} & 0.48 & 0.28 \\
\midrule
\multicolumn{6}{l}{\textit{Reasoning Mode}} \\
Claude S4.5 (Think) & 0.71 & 0.62 & 0.66 & 0.44 & 0.34 \\
DS Reasoner         & 0.70 & 0.64 & 0.55 & 0.48 & 0.33 \\
Qwen3 Max Reason    & \textbf{0.75} & 0.64 & 0.68 & \textbf{0.49} & 0.26 \\
Gemini 3 Pro Low    & 0.66 & 0.58 & 0.64 & 0.48 & 0.31 \\
Gemini 3 Pro High   & 0.66 & 0.63 & 0.64 & 0.48 & 0.33 \\
\bottomrule
\end{tabular}
\end{table}

\section{Tool Use Data Generation Prompt}
\label{app:tool_use_generation_prompt}

\subsection{Prompt for TravelPlanner}
\begin{lstlisting}[breaklines=true, breakatwhitespace=true, frame=single, backgroundcolor=\color{gray!10}]
### Source Context
This is a travel planning scenario in the United States.
Origin City: {origin}
Destination Cities: {', '.join(destinations) if destinations else 'Various'}
Trip Dates: {', '.join(sorted(dates)) if dates else 'Not specified'}
Available Cities in Data: {', '.join(cities)}

### Your Task

**Generate exactly 5 Single-Step Tool Call Samples**
Generate diverse single-step tool calls covering different tool types:
- Flight search between cities
- Accommodation search in destination cities
- Attraction search in destination cities
- Restaurant search in destination cities
- Distance/transportation queries between cities

Reference single-step query examples (generate similar ones based on the trip context):
{single_step_hints}

### Available Tools
{TRAVELPLANNER_TOOLS_DESC}

### Complete Examples (Query + Tool Call)

Example 1 - FlightSearch:
- Query: "Find flights from Washington to Myrtle Beach on March 13, 2022."
- Tool: FlightSearch("Washington", "Myrtle Beach", "2022-03-13")

Example 2 - AccommodationSearch:
- Query: "Search for hotels in Atlanta."
- Tool: AccommodationSearch("Atlanta")

Example 3 - AttractionSearch:
- Query: "What attractions can I visit in Tucson?"
- Tool: AttractionSearch("Tucson")

Example 4 - RestaurantSearch:
- Query: "Find restaurants in Myrtle Beach."
- Tool: RestaurantSearch("Myrtle Beach")

Example 5 - DistanceMatrix:
- Query: "How long does it take to drive from Washington to Myrtle Beach?"
- Tool: DistanceMatrix("Washington", "Myrtle Beach", "self-driving")

### Output Format (JSON ONLY)
Return a single JSON object with exactly 5 samples:

**Field Descriptions:**
- `query_en`: Natural language query in English (must include all specific parameter values like city names, dates)
- `query_zh`: Natural language query in Chinese (must include all specific parameter values)
- `tool_name`: The exact function name to call (e.g., "FlightSearch", "AccommodationSearch")
- `parameters`: A JSON object containing all required parameters for the tool

{example_output}

### Important Rules
1. **Use Cities from Context**: Generated queries MUST use the actual cities from the trip context: {', '.join(cities)}
2. **Use Dates from Context**: For flight queries, use the actual dates: {', '.join(sorted(dates)) if dates else 'use reasonable dates'}
3. **Be Specific**: Queries MUST explicitly specify ALL parameter values (city names, dates, etc.)
4. **Maintain Relevance**: All queries should be related to planning a trip between the given cities
5. **Cover Tool Variety**: Try to use different tool types (don't just use FlightSearch for all 5)
6. Use English city names exactly as they appear in the data
7. Generate natural, diverse queries, avoid repetition
\end{lstlisting}

\subsection{Prompt for TripCraft}
\begin{lstlisting}[breaklines=true, breakatwhitespace=true, frame=single, backgroundcolor=\color{gray!10}]
### Source Context
This is a travel planning scenario in the United States.
Original Query: "{query}"
Origin City: {origin}
Destination City: {destination}
Trip Duration: {days} days
Trip Dates: {date_range_str}
Number of People: {people}
Budget: ${budget}

### Your Task

**Generate exactly 5 Single-Step Tool Call Samples**
Generate diverse single-step tool calls covering different tool types:
- Flight search between cities
- Accommodation search in destination city
- Attraction search in destination city
- Restaurant search in destination city
- Event search in destination city (with date range)
- Distance/transportation queries between cities

Reference single-step query examples (generate similar ones based on the trip context):
{single_step_hints}

### Available Tools
{TRIPCRAFT_TOOLS_DESC}

### Complete Examples (Query + Tool Call)

Example 1 - Flights:
- Query: "Find flights from Savannah to Baltimore on November 18, 2024."
- Tool: Flights("Savannah", "Baltimore", "2024-11-18")

Example 2 - Accommodations:
- Query: "Search for hotels in Baltimore."
- Tool: Accommodations("Baltimore")

Example 3 - Attractions:
- Query: "What attractions can I visit in Baltimore?"
- Tool: Attractions("Baltimore")

Example 4 - Restaurants:
- Query: "Find restaurants in Baltimore."
- Tool: Restaurants("Baltimore")

Example 5 - Events:
- Query: "Find events in Baltimore from November 18 to November 20, 2024."
- Tool: Events("Baltimore", ["2024-11-18", "2024-11-20"])

Example 6 - GoogleDistanceMatrix:
- Query: "How long does it take to drive from Savannah to Baltimore?"
- Tool: GoogleDistanceMatrix("Savannah", "Baltimore", "driving")

### Output Format (JSON ONLY)
Return a single JSON object with exactly 5 samples:

**Field Descriptions:**
- `query_en`: Natural language query in English (must include all specific parameter values like city names, dates)
- `query_zh`: Natural language query in Chinese (must include all specific parameter values)
- `tool_name`: The exact class name to call (e.g., "Flights", "Accommodations", "Events", "GoogleDistanceMatrix")
- `parameters`: A JSON object containing all required parameters for the tool

{example_output}

### Important Rules
1. **Use Cities from Context**: Generated queries MUST use the actual cities: {origin} and {destination}
2. **Use Dates from Context**: For flight/event queries, use the actual dates: {date_range_str}
3. **Be Specific**: Queries MUST explicitly specify ALL parameter values (city names, dates, etc.)
4. **Maintain Relevance**: All queries should be related to planning a trip from {origin} to {destination}
5. **Cover Tool Variety**: Try to use different tool types (Flights, Accommodations, Attractions, Restaurants, Events, GoogleDistanceMatrix)
6. Use English city names exactly as they appear
7. Generate natural, diverse queries, avoid repetition
\end{lstlisting}

\subsection{Prompt for ChinaTravel}
\begin{lstlisting}[breaklines=true, breakatwhitespace=true, frame=single, backgroundcolor=\color{gray!10}]
### Source Context
User Request: "{user_request}"
Origin City: {start_city}
Destination City: {target_city}
Trip Duration: {days} day(s)
Number of People: {people}

### Budget Allocation Rules (VERY IMPORTANT!)
{f"Total Budget: {budget} RMB " if budget else "Total Budget: Not specified, extract it from user request if mentioned."}
- Trip Duration: {days} day(s)
- Number of People: {people}

**Allocate budget reasonably** across different expenses:
- Hotel: Reasonable per-night price (NOT the total budget)
- Restaurant: Reasonable per-meal price (50-150 RMB typical)
- Transportation: Estimate based on distance
- Attractions: Reasonable ticket price filtering

**WRONG Example**: Budget 1500 RMB, generate `accommodations_select("Hangzhou", "price", "lambda x: x <= 1500")`
**CORRECT Example**: Budget 1500 RMB, generate `accommodations_select("Hangzhou", "price", "lambda x: x <= 400")`

Filter conditions must be reasonable, ensuring all expenses combined do not exceed the total budget!

### Your Task

**Generate exactly 5 Single-Step Tool Call Samples**
Generate diverse single-step tool calls covering different tool types:
- Attraction search/filter
- Accommodation search/filter
- Restaurant search/filter
- Intercity transport query
- City transport query

Reference single-step query examples (generate similar ones based on user request):
{single_step_hints}

### Available Tools
{WORLD_ENV_TOOLS_DESC}

### Complete Examples (Query + Tool Call)

Example 1 - attractions_select:
- Query: "I want to visit attractions in Hangzhou with rating at least 4.5."
- Tool: attractions_select("Hangzhou", "rating", "lambda x: x >= 4.5")

Example 2 - accommodations_select:
- Query: "Find me hotels in Hangzhou with price no more than 400 RMB per night."
- Tool: accommodations_select("Hangzhou", "price", "lambda x: x <= 400")

Example 3 - restaurants_nearby:
- Query: "Recommend 5 restaurants within 2km of West Lake in Hangzhou."
- Tool: restaurants_nearby("Hangzhou", "West Lake", 5, 2)

Example 4 - intercity_transport_select (BETWEEN DIFFERENT CITIES by train/airplane):
- Query: "I need a train from Shanghai to Hangzhou departing after 8 AM."
- Tool: intercity_transport_select("Shanghai", "Hangzhou", "train", "08:00")
- NOTE: Shanghai and Hangzhou are DIFFERENT cities, so use intercity_transport_select

Example 5 - goto (WITHIN SAME CITY by taxi/metro/walk):
- Query: "How do I take a taxi from West Lake to Lingyin Temple at 10 AM in Hangzhou?"
- Tool: goto("Hangzhou", "West Lake", "Lingyin Temple", "10:00", "taxi")
- NOTE: West Lake and Lingyin Temple are both in Hangzhou (SAME city), so use goto

Example 6 - goto (metro within city):
- Query: "Take the metro from Hangzhou East Station to West Lake at 9:30 AM."
- Tool: goto("Hangzhou", "Hangzhou East Station", "West Lake", "09:30", "metro")
- NOTE: Both locations are in Hangzhou, using metro (local transport), so use goto

### Output Format (JSON ONLY)
Return a single JSON object with exactly 5 samples:

**Field Descriptions:**
- `query_en`: Natural language query in English (must include all specific parameter values)
- `query_zh`: Natural language query in Chinese (must match query_en semantically)
- `tool_name`: The exact function name to call (e.g., "attractions_select", "goto")
- `parameters`: A JSON object containing all required parameters for the tool

{example_output}

### Important Rules
1. **Maintain Relevance**: Generated queries MUST be related to the user's original request!
   - Destination city is {target_city}, all attraction/hotel/restaurant queries should be in {target_city}
   - Origin city is {start_city}, intercity transport should be {start_city} -> {target_city}
   - Do NOT generate unrelated cities or locations (e.g., if user goes to Hangzhou, don't query Beijing hotels)

2. **Be Specific with Parameters**: Queries MUST explicitly specify ALL parameter values, NO vague expressions!
   - WRONG: "Find restaurants nearby" (vague - what location? how far?)
   - CORRECT: "Find 5 restaurants within 2km of West Lake"
   - WRONG: "Find hotels around 300 RMB" (vague - what range?)
   - CORRECT: "Find hotels under 300 RMB per night"
   - WRONG: "Find highly rated attractions" (vague - what rating?)
   - CORRECT: "Find attractions with rating 4.5 or above"
   - Always include: exact numbers, specific locations, clear comparison operators (>=, <=, ==, etc.)

3. **Comparison Operators - VERY IMPORTANT**:
   - AVOID ambiguous words like "under" / "less than" - they are confusing!
   - USE CLEAR expressions instead:
     - "no more than X" / "at most X" / "X or less" -> use `<= X`
     - "at least X" / "X or above" / "X or higher" -> use `>= X`
     - "more than X" / "above X" -> use `> X`
   - For range queries like "between 50 and 100", use `lambda x: 50 <= x <= 100` format
   - EXAMPLES of GOOD vs BAD expressions:
     - BAD: "under 300 RMB" (ambiguous: < or <=?)
     - GOOD: "no more than 300 RMB" (clear: <=)
     - GOOD: "at most 300 RMB" (clear: <=)
     - GOOD: "300 RMB or less" (clear: <=)

5. **SINGLE FILTER CONDITION ONLY - CRITICAL!!!**:
   - Each *_select tool call MUST filter on ONLY ONE field (key)!
   - DO NOT generate queries with multiple filter conditions on different fields!
   - BAD EXAMPLES (NEVER generate these):
     - "Find attractions with ticket price under 80 RMB AND rating above 4.3" (2 different keys: price + rating)
     - "Find hotels under 300 RMB with rating above 4.0" (2 different keys: price + rating)
     - "Find restaurants with rating above 4.5 and average cost under 100" (2 different keys)
   - GOOD EXAMPLES (single condition only):
     - "Find attractions with rating at least 4.5" (only rating)
     - "Find hotels no more than 300 RMB per night" (only price)
     - "Find restaurants with rating 4.0 or above" (only rating)
   - If you need multiple conditions, generate separate tool calls (multi-step scenario)

4. **TRANSPORTATION TOOL SELECTION - MOST CRITICAL!!!**:

   #### goto (INTRA-CITY / LOCAL transport)
   - **USE FOR**: Moving between locations WITHIN THE SAME CITY
   - **Transport types**: taxi, metro, walk (LOCAL transport only!)
   - **Key indicators**: "take a taxi to...", "by metro to...", "walk to...", "how to get from A to B" (same city)
   - **Examples**:
     - "Take a taxi from West Lake to Lingyin Temple" -> goto("Hangzhou", "West Lake", "Lingyin Temple", "10:00", "taxi")
     - "How to get from the train station to the hotel by metro" -> goto(city, "Train Station", "Hotel", time, "metro")
     - "Walk from the museum to the restaurant" -> goto(city, "Museum", "Restaurant", time, "walk")

   #### intercity_transport_select (INTER-CITY / LONG-DISTANCE transport)
   - **USE FOR**: Traveling BETWEEN DIFFERENT CITIES
   - **Transport types**: train, airplane (LONG-DISTANCE transport only!)
   - **Key indicators**: "from CityA to CityB", "fly to...", "take a train to another city"
   - **Examples**:
     - "Train from Shanghai to Hangzhou" -> intercity_transport_select("Shanghai", "Hangzhou", "train", "08:00")
     - "Flight from Beijing to Guangzhou" -> intercity_transport_select("Beijing", "Guangzhou", "airplane", "10:00")

   #### DECISION RULE:
   | Scenario | Tool | Why |
   |----------|------|-----|
   | Taxi from hotel to attraction (same city) | goto | Same city, local transport |
   | Metro from train station to scenic spot | goto | Same city, local transport |
   | Train from Shanghai to Hangzhou | intercity_transport_select | Different cities |
   | "How to get from A to B" within city | goto | Same city movement |

   **NEVER use intercity_transport_select for taxi/metro/walk!**
   **NEVER use goto for train/airplane between cities!**

5. **topk Parameter**: When query specifies a number (like "5 restaurants" / "3 hotels"), use that exact number for topk
   - WRONG: "Find 5 restaurants" -> restaurants_nearby(..., topk=10, ...)
   - CORRECT: "Find 5 restaurants" -> restaurants_nearby(..., topk=5, ...)

6. All `func` parameters must be valid Python lambda strings
7. Use English city names: {target_city}
8. If there's a budget constraint, reflect it in accommodations_select and restaurants_select with reasonable per-item prices
9. Generate natural, diverse queries, avoid repetition
10. Cover as many tool types as possible
\end{lstlisting}

\section{LLM as a Judge Prompt}
\label{app:llm_as_a_judge_prompt}
\FloatBarrier
\begin{lstlisting}[breaklines=true, breakatwhitespace=true, frame=single, backgroundcolor=\color{gray!10}]
# The prompt used for the LLM as a judge is as follows:

You are a formal "constraint equivalence judge" for DSL constraint extraction.
Input (do NOT include UID in your output):
- DSL documentation / function semantics:
{dsl_block}

- GOLD constraints (array of strings):
{gold_json}

- PRED constraints (array of strings):
{pred_json}

Task (precise):
1. For each PRED constraint, determine whether it is functionally equivalent to any single GOLD constraint.
   - "Functionally equivalent" means: given the DSL semantics above and for any reasonable plan, the boolean truth value of the PRED equals that of the GOLD expression it is said to match.
   - Allow matching across minor syntactic differences (e.g., `people_count(plan)==2` vs `result=(people_count(plan)==2)`), but do not match if semantics differ.
2. Multiple PREDs may map to the same GOLD. That is allowed.
3. Any GOLD not matched by any PRED counts as a miss (FN). Any PRED that matches no GOLD counts as a false positive (FP). Count true positives (TP) as PREDs that matched a GOLD.

Important output rules (strict):
- Do NOT output any chain-of-thought or internal deliberation. Do NOT reveal step-by-step reasoning.
- Provide only concise, non-sensitive justifications (1 short sentence, maximum 25 words) per prediction as the value `brief_reason`.
- Do NOT include the UID in your output. The calling program will attach the UID later.
- Output exactly one JSON object (and nothing else). The JSON must match the schema below exactly.

Required JSON schema (produce this exact structure):
{{
  "matches": [
    {{
      "pred": "<the pred constraint string exactly as given>",
      "matched_gold_idx": <integer index of the matched gold constraint, 0-based, or null if no match>,
      "brief_reason": "<one short, non-sensitive sentence explaining the match or why no match>"
    }}
    ... (one entry per pred in the same order as input)
  ],
  "tp": <integer>,
  "fp": <integer>,
  "fn": <integer>,
  "tn": 0,
  "summary": "<one short sentence summarizing overall judgment>"
}}

Additional notes (do not output these):
- matched_gold_idx must refer to an index in the provided GOLD array (0-based). Use null if the pred matches no gold.
- Count tp as the number of matches where matched_gold_idx is not null. Count fp as the number of matches where matched_gold_idx is null. Compute fn as (number of gold constraints) minus (number of distinct matched gold indices).
- Set tn to 0 (the negative class is not defined for this task).
- Keep each brief_reason strictly short (<=25 words). The final summary should be one sentence.

Begin and output only the required JSON object.
\end{lstlisting}
\FloatBarrier


\end{document}